\pdfoutput=1
\documentclass{article}

     \PassOptionsToPackage{numbers, compress}{natbib}

    \usepackage[final]{neurips_2020}

\usepackage[utf8]{inputenc} 
\usepackage[T1]{fontenc}    
\usepackage{hyperref}       
\usepackage{url}            
\usepackage{booktabs}       
\usepackage{amsfonts}       
\usepackage{nicefrac}       
\usepackage{microtype}      
\usepackage{color}
\usepackage{graphicx}
\usepackage{algorithm}
\usepackage{algpseudocode}
\usepackage{array}
\usepackage{csquotes}
\title{Explaining Conditions for Reinforcement Learning Behaviors from Real and Imagined Data}

\author{%
    Aastha Acharya \\
    Aerospace Engineering Sciences\\
    University of Colorado, Boulder \\
    The Charles Stark Draper Laboratory, Inc.\\
    \texttt{aastha.acharya@colorado.edu} \\
    \And
    Rebecca Russell \\
    The Charles Stark Draper Laboratory, Inc. \\
    \texttt{rrussell@draper.com} \\
    \And
    Nisar R. Ahmed \\
    Aerospace Engineering Sciences\\
    University of Colorado, Boulder \\
    \texttt{nisar.ahmed@colorado.edu}
}

\begin{document}

\maketitle

\begin{abstract}
The deployment of reinforcement learning (RL) in the real world comes with challenges in calibrating user trust and expectations. As a step toward developing RL systems that are able to communicate their competencies, we present a method of generating human-interpretable abstract behavior models that identify the experiential conditions leading to different task execution strategies and outcomes. Our approach consists of extracting  experiential features from state representations, abstracting strategy descriptors from trajectories, and training an interpretable decision tree that identifies the conditions most predictive of different RL behaviors. We demonstrate our method on trajectory data generated from interactions with the environment and on imagined trajectory data that comes from a trained probabilistic world model in a model-based RL setting. 

\end{abstract}

\section{Introduction}

Reinforcement learning (RL) continues to be a promising field with demonstrated success in a wide variety of challenging tasks from Atari \cite{mnih2013playing, kaiser2019modelbased} to robotics \cite{Finn_2017, TobinFRSZA17, MBRLRobotics}. However, RL deployment in most real world applications still requires significant advances in several areas within the field. Dulac-Arnold, et al. \cite{dulacarnold2019challenges} list nine challenges that encompass problems ranging from the sparsity of training data to the challenge of designing appropriate reward functions. We argue that the challenge of RL explainability and user trust is the most fundamental for safe real world deployment of these systems. User trust in RL goes beyond just policy explainability, but to designing RL systems that are able to communicate their tasking competencies: their relevant experience, strategies, and self-confidence. Competency communication is particularly essential for early stage RL systems used in real world settings, which may have surprising behaviors that are mismatched with user expectations. We take our first steps towards developing competency-communicating RL by developing explainable abstract behavior models of RL agents.

RL agents that communicate their competencies must not only be able to convey what they intend to do in specific scenarios, but also communicate their general behavior and the conditional dependencies leading to different strategies and outcomes. To achieve this, we abstract important experiential features and associated task strategies and outcomes from a trained RL agent's trajectory data. This abstraction is performed at two levels: at individual time steps within the trajectory and based on overall outcome for a given trajectory sequence. In the model-based RL framework, we are able to use \emph{imagined} trajectory rollouts in a probabilistic world model to make these same inferences without requiring any additional interaction with the environment.

\textbf{Related work } Approaches to explainable RL tend to focus on building interpretable policies~\cite{shu2017hierarchical, verma2018programmatically} or explaining policies through visualizations and examples \cite{coppens2019distilling, Sequeira_2020, huangtrust}. Using abstract behavior models to describe and explain RL agent behavior in structured language has been explored in contexts with simple behavior conditions. Hayes and Shah~\cite{hayes2017improving} model and explain robot control policies by representing state spaces in terms of communicable predicates and finding the disjunctive normal form representations of conditions for actions. Van der Waa, et al. \cite{waa2018contrastive} find contrastive explanations for RL agent policies based on expected outcomes compared to user-suggested policies with a user-interpretable MDP. These approaches are based on explaining policies in terms of actions and rely on predetermined state predicates, preventing them from scaling to complex RL tasks. Our approach instead combinatorially builds experiential features that are used to identify the conditions leading to different behaviors using an interpretable decision tree. 

\section{Abstract Behavior Model}\label{approach}

\begin{figure}
  \centering
  \includegraphics[scale=0.36]{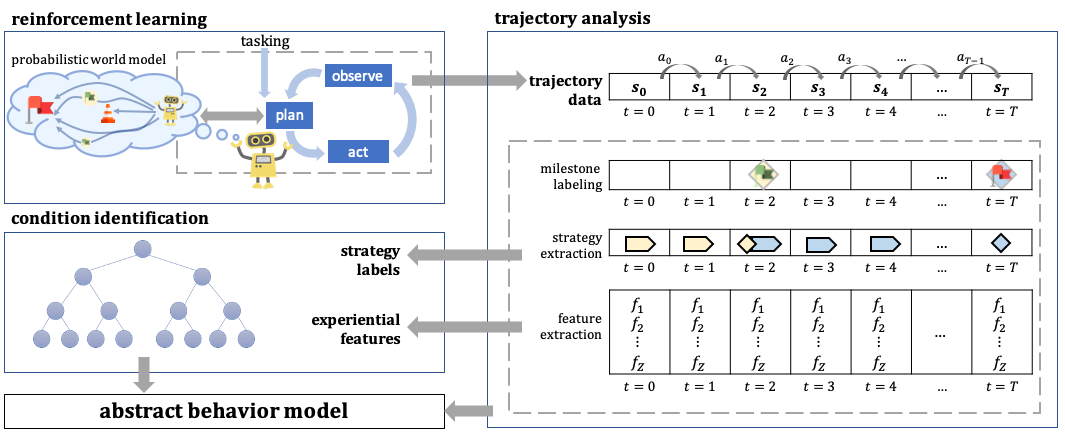}
  \caption{Our pipeline analyzes real or imagined RL trajectory data to identify the experiential conditions that lead to different strategies.}
  \label{frmwork}
\end{figure}

Our work focuses on developing an abstract behavior model in which to understand and contextualize RL competency. This model consists of two major components: \emph{behaviors} (a combination of agent \emph{strategies} and \emph{outcomes}) and their corresponding experiential \emph{conditions}. We show how to process a set of trajectories from a trained RL agent to generate human-interpretable experiential features and strategy labels. We then train a decision tree with trajectory data in this feature representation to identify conditions that are associated with the different strategy labels or outcomes, where a condition represents a conjunction of relevant experiential feature predicates. Collectively, the conditions from the decision tree and strategy labels form an abstract behavior model that allows us to present RL experiences and strategies at a higher level from raw trajectory data. This model can form the communication basis for a full competency assessment of the RL system. We show how the results in their current form explain the behavior of an agent in a human interpretable manner. Full details on our methodology are presented in the following subsections with an accompanying graphic shown in Figure \ref{frmwork}. 

\subsection{Preliminaries and Notation}\label{approach:RL}

Reinforcement learning \cite{Sutton1998} is formalized as a Markov Decision Process (MDP) consisting of a set of states $\mathcal{S}$, set of actions $\mathcal{A}$, a reward function $r(s_t,a_t)$ that maps the current state $s_t$ and action $a_t$ to a reward value, and a transition dynamics function $p(s_{t+1}|s_t, a_t)$ that provides the distribution over the next state $s_{t+1}$ given the current state $s_t$ and action $a_t$. A policy $\pi$ in RL provides an action from any state such that $a_t \sim \pi(s_t)$, and its execution results in a \textbf{trajectory}  $\mathcal{D} = \{(s_t, a_t, s_{t+1})\}_{t=1:T}$. 

We extract the behavior of the RL agent from a trajectory dataset by first analyzing states at individual times within each trajectory. We expand state representations into interpretable \textbf{experiential features} which contain detailed contextual information on entities represented within states, the relationships between these entities, and comparisons between the relationships. These experiential features represent factors that may impact an agent's behavior, and can be in either discrete or continuous form. We use $f_t = \mathbf{f}(s_t)$ to denote the experiential feature vector at time $t$. 

To collect information on the strategies represented in trajectory data, we rely on trajectory \textbf{milestones}. Milestones represent notable accomplishments of the RL agent within a trajectory and are defined by an expert based on their domain knowledge of the task. They are a pre-generated list of classifiers that take in the current and the previous states and output a binary value based on a match to our criteria for that milestone. These criteria may be thresholds on individual states, equality of two more states, or time-dependent change in states. If the reward function is available, the states and their criteria from within the reward function may also be used as milestones. The resulting $M$ milestone classifiers are represented as $\mathcal{M} = \{m_j\}_{j=1:M}$. 

Once we extract the milestones within each trajectory, we generate \textbf{strategy labels}, denoted $\mathcal{Y} = \{y_{t}\}_{t=1:T}$, at each time $t$ in the trajectory. These labels directly correspond to the next milestone reached because we assume that all actions leading up to milestone represent an agent's strategy to get there. We use the term \enquote{strategy} loosely here because it is possible for the milestones to represent states that do not factor into the reward function and therefore the agent is not directly optimizing for. 

Building upon these components, a \textbf{condition} in our setting means a conjunction of experiential feature predicates that correlates with a given strategy label. We use the trajectory \textbf{outcome} to mean the categorization of a trajectory as either a success or a failure. Finally, the \textbf{behavior} of an agent more broadly represents both the strategy and the outcome for a given trajectory. 

\subsection{Trajectory Collection from Real and Imagined World}

Our method is very general as the trajectory dataset used to derive the abstract behavior model can come from any source, either model-free or model-based RL, and from any policy, either optimal or otherwise. While typical trajectory data is assumed to be produced from direct interactions with the environment, our method is not restricted in that regard and can be easily applied to imagined trajectory datasets.

In the model-based RL setting, we learn a model of the environment by learning the transition dynamics function $\hat{p}(s_{t+1}|s_t, a_t)$ that is an approximate to the true environment $p(s_{t+1}|s_t, a_t)$. One of the demonstrated benefits of model-based RL is the agent's ability to use this trained world model to imagine the real environment \cite{ha2018worldmodels, hafner2018learning}. Furthermore, a probabilistic world model \cite{chua2018deep, malik2019calibrated} can generate diverse imagined trajectory data. We show that our method can be directly applied to this data without any alterations. To do this, we maintain the same policy as with the real environment, and have several options to initialize the first state: 1) randomly, 2) by hand, or 3) using observations from the environment. Randomly sampling the first state allows us to generate diverse trajectory data that cover a wide range of possible experiences, with the quality of the trained model being the limiting factor. Selection of the first state by hand or receiving an observation from the environment allows us to target the statistics of our behavior model and perform multiple rollouts from one state to thoroughly explore the variety of trajectories that can result due to the stochasticity of the environment. Once we have an initial state and an action, we input them into the probabilistic world model to receive a prediction of the next state distribution. Then, we set our next state by sampling from the probabilistic prediction such that $\hat{s}_{t+1} \sim \hat{p}(s_{t+1}|s_t, a_t)$. The trajectories in our imagined dataset take the form $\{\hat{s}_{t}, a_{t}, \hat{s}_{(t+1)}\}_{t=1:T}$. This method of generating the data gives us the ability to speak to the competency of the model-based RL agent without having to collect data from the environment. Additionally, using an abstract behavior model to compare the data produced from real and imagined environment may also provide insight into the differences between the two and the overall quality of the trained model, which is another component of competency for model-based RL.

\subsection{Trajectory Analysis}\label{approach:strategy}

Our trajectory analysis consists of two parts: trajectory abstraction and strategy extraction. The trajectory abstraction process produces information on the experiential features at a given time and the milestones reached during an episode. The full procedural details are shown in Algorithm \ref{alg:traj}. 

Once the trajectory milestone information (denoted $\mathcal{X}$) and the experiential features (denoted $F$), have been extracted, they are used as an input to the strategy extraction procedure described in the second half of Algorithm \ref{alg:traj}. For this step, we assign a strategy label at each time by looking ahead to the next milestone reached. This may be as simple as assigning strategy label $j$ if the next milestone accomplished is $m_j$ or may be more complex if we choose to consider space representations and the experiential features. We provide an example of both approaches in our experiments in Section \ref{experiments}. 

\begin{algorithm} 
	\caption{Trajectory Analysis} 
	\label{alg:traj}
	\begin{algorithmic}[1]
	    \Procedure{Trajectory Abstraction}{$ \mathcal{D} = \{(s_{t},a_{t},s_{t+1})\}_{t=1:T}, \mathcal{M}=\{m_j\}_{j=1:M}$} \label{alg:proc1}
		    \For {time $t=1,2,\ldots,T$}
	            \State Extract experiential features $f_t \gets \mathbf{f}(s_t)$
    		    \For{milestone classifier $j=1,2,\ldots,M$}
		            \State Assign $x_{tj} \gets m_j(s_{t}, s_{t+1})$ for satisfaction of milestone criteria
                \EndFor
            \EndFor
    		\State \textbf{return} experiential features $F \gets \{f_{t}\}_{t=1:T}$ 
            \State \textbf{return} milestone list $\mathcal{X} \gets \{x_{tj}\}_{t=1:T, j=1:M} $
    	\EndProcedure
    	\State
    	\Procedure{Strategy Extraction}{$F=\{f_{t}\}_{t=1:T}, \mathcal{X}=\{x_{tj}\}_{t=1:T, j=1:M}$} \label{alg:proc2}
	        \State Extract total number of times a milestone is reached $N \gets \sum_{t=0}^T{\max(\{x_{tj}\}_{j=1:M})}$
	        \State Extract times when milestones reached $t^* \gets \{t^*_{n}\}_{n=1:N}$
	        \For {time $t=1, 2,\ldots,T$}
	            \For {milestones $n=1,2\ldots, N$}
	                \If {$t\geq t^*_{n-1}$ and $t<t^*_{n}$}
	                    \State Assign strategy $y_{t}$ using ($s_{t}, f_{t}$) relative to ($x_{t^*_m}, f_{t^*_m}$) 
	                \EndIf
	            \EndFor
	        \EndFor
    	    \State \textbf{return} strategy labels $\mathcal{Y} \gets \{y_{t}\}_{t=1:T}$
    	\EndProcedure
	\end{algorithmic} 
\end{algorithm}

\subsection{Condition Identification} \label{approach:cond}

The experiential features $F$ and strategy labels $\mathcal{Y}$ that are produced from the trajectory abstraction process are used to perform condition identification and generate the abstract behavior model. We use decision tree learning to greedily partition the dataset by individual experiential features to best predict the strategy of our RL agent. Furthermore, we can fully exploit the benefits of a decision tree, including its ease of interpretability and structured form, to produce information on the hierarchy and importance of experiential features. We limit the complexity of the decision tree by placing a minimum required Gini impurity decrease for each additional split and a maximum tree depth. 

We have flexibility in generating our desired level of granularity in identifying the conditions that lead to outcomes and strategies. For example, if we are only interested in the outcome of an episode given the observations at particular states or times, it is possible to only provide the final outcome and the relevant feature information to identify the conditions that lead to a success or a failure. However, if there are sub-tasks provided to the RL agent or if we are interested in sub-strategies, the experiential features and strategy labels at all times may be used as training data.

\section{Experiments} \label{experiments}

To study and demonstrate our approach, we design a simple grid world environment where an agent's objective is to reach the goal while collecting as many fuel pods as possible and avoiding the adversary that is actively, and stochastically, chasing it. Using a model-based RL agent, we generate abstract behavior models derived from both real and imagined trajectory data. Our identified conditions and strategies are presented in a decision tree that helps to explain the agent's conditional behavior. The results shown take us a step closer to our overarching objective of building a system capable of assessing and communicating the competency of a RL agent.

\subsection{Experimental Design} \label{exp:env}

\textbf{Environmental Setup } The environment consists of an agent, an adversary, a goal, and three fuel pods that are all scattered randomly throughout an $8\times8$ map.  An episode is considered to be successful if the agent reaches the goal without being caught by the adversary. There are also sub-tasks for the agent, such as collecting the fuel pods, which is a desired, but not a required, objective. In addition to seeing  a new map at each initial condition, additional complexity is introduced into the environment by allowing randomness in the adversary's behavior while it is chasing the agent. Designing the environment in this way allows sub-strategies to emerge that we can analyze at a time-varying level. The state space consists of four $8\times8$ channels consisting of location information for the agent, adversary, goal, and fuels, respectively. The actions for the agent are represented in a $4$-dimensional categorical space, corresponding to up, down, right, and left. 
\begin{figure}
  \centering
  \includegraphics[scale=0.37]{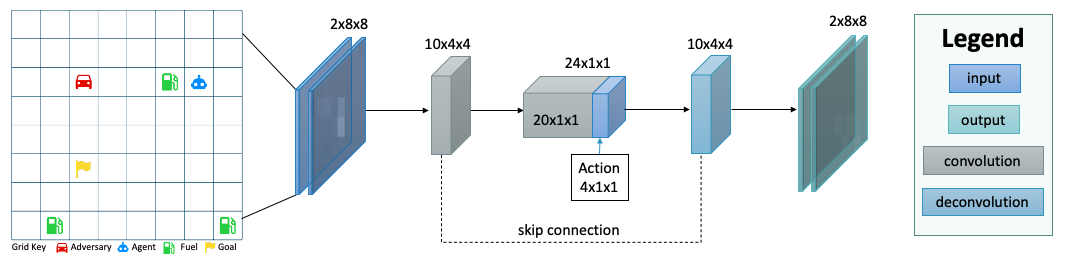}
  \caption{Grid world environment (left) and the world model neural network architecture (right)}
  \label{env}
\end{figure}

\textbf{Model-Based Reinforcement Learning } Our experiments focus on using model-based RL \cite{moerl2020modelbased, wang2019benchmarking} due to its sample efficiency \cite{Nagabandi_2018}, multi-tasking flexibility \cite{ebert2018visual}, and ability to imagine trajectory data using the learned world model \cite{ha2018worldmodels, hafner2018learning}. Our goal in model-based RL is to learn an approximate world model $\hat{p}(s_{t+1}|s_t, a_t)$ of the real world dynamics $p(s_{t+1}|s_t, a_t)$. More specifically, we are learning the agent's and the adversary's next state given their current states and agent action. Since the fuel and goal locations do not change during an episode, we do not include these states in the model learning. We define a reward function that assigns a positive reward to the agent reaching each of the three fuels and the one goal, with the goal reward being slightly higher value than the fuel rewards. To incentivize the agent to collect the fuels before going to the goal, we dynamically change the goal reward to be higher only once all the fuel pods have been seized. Additionally, capture by the adversary is represented in the reward function with a penalty whose absolute value is equal to that of reaching the goal. 

\textbf{Model Learning and Planner } We use a U-Net style convolutional neural network (CNN), shown in Figure \ref{env}, as our probabilistic world model. The network takes in the binary two-channel position information on the agent and the adversary of size $2\times8\times8$. Following two convolution operations that reduce the network down to state of $20 \times 1 \times 1$, the action input, represented as $4\times1\times1$ in a one-hot encoded form, is concatenated to produce a $24 \times 1 \times 1$ sized layer. This is followed by two layers of deconvolution, resulting in a prediction of the next state for both the agent and the adversary. We use model predictive control (MPC) with a discrete tree-based planner that aggregates all four actions from any given state and generates sequences of possible trajectories over the provided planning horizon. These trajectories are then ranked using the reward function in order to pick the best sequence and the associated actions. The re-planning occurs at every time step, and we only apply the first action from the best sequence at that given time. 

\textbf{Imagined Trajectories } We use the trained probabilistic world model to generate additional trajectory data to produce another abstract behavior model purely from the imagined trajectory rollouts. The probabilistic world model outputs a categorical probability distribution over each channel, including epistemic uncertainty through model ensembling~\cite{lakshminarayanan2017simple}. We randomly initialize the first state and use our planner and world model to select an action. Then, inputting the state and action into the probabilistic world model, we sample $\hat{s}_{t+1} \sim \hat{p}(s_{t+1}|s_t, a_t)$ from the predicted probability distribution to get the next estimate for the agent's and adversary's state. This recursive process terminates once the agent reaches the goal, is caught by the adversary, or if the time given to complete the task has elapsed, all within the imagined world. 

\subsection{Trajectory Processing}\label{trajpro}

Given the trajectory data, we extract the experiential features from the states at each time step. These experiential features include the current location information for all the entities in the environment, the distances between them, and binary values comparing these distances. For example, three of the experiential features are: agent's $y$ position, distance between adversary and the goal at the given time, and the binary value of whether the distance between the agent and the goal is greater than the distance between the adversary and the goal. We also categorize the fuel according to its relative distance to all the other components, so our experiential features contain information on the fuel closest to the agent, the fuel closest to the goal, and the fuel furthest from the adversary at any given time. In total, we have $132$ experience features that we extract from the observed states at each time. 

To define a list of milestones, we rely on our domain knowledge of the task and the reward function. The most impactful states are when the agent reaches any of the fuels, the goal, or is caught by the adversary, so we label these as our milestones. We create our strategy labels for each time based on the next milestone. If the next milestone reached is \enquote{agent at goal,} \enquote{agent caught,} or \enquote{unterminated,} we directly assign that strategy label. However, if the next reached milestone is a fuel location, we label the strategy based on the location of that fuel relative to the agent, goal, and adversary at each time. Hence, the strategies leading up to the fuel milestones may change before the milestone is reached (e.g., from showing that the agent is going for the fuel closest to the goal to showing that the agent is going for a the fuel that is closest to itself). As a result, we end up with a total of seven local strategies for this task: {1) agent at goal, 2) agent caught, 3) unterminated episode, 4) agent at unlabeled fuel, 5) agent at fuel closest to agent, 6) agent at fuel closest to goal, and 7) agent at fuel furthest from adversary}. 

The agent's behavior is modeled at two levels: determining the outcome of an episode and predicting the local strategy. For modeling the outcome, we only use the experiential features from the initial state to relate how our varying initial conditions affect the final outcome. For modeling the strategy, we predict the local strategy based on the time-varying experiential features. For the strategy modeling, we exclude failed episodes in order to derive strategies for successful trajectory runs only. Both levels of the abstract behavior model are based on a independently-trained interpretable decision tree.

\subsection{Results} \label{exp:results}

\begin{figure}
  \centering
  \includegraphics[scale=0.42]{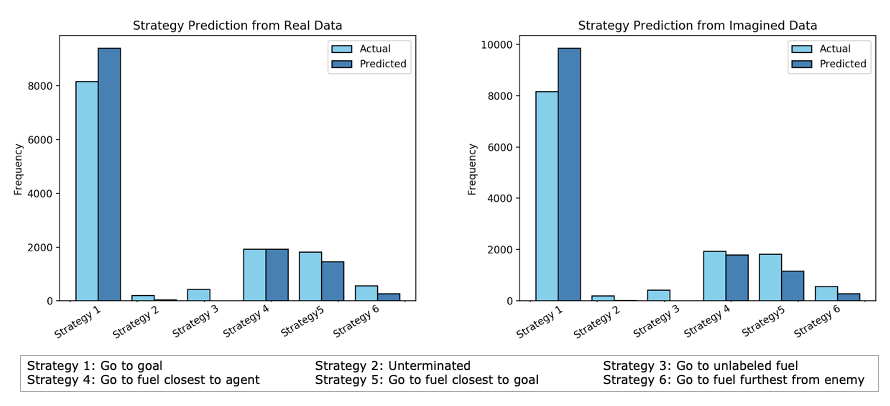}
  \vspace{+.25cm}
  \caption{Actual and predicted strategy frequencies in both real (left) and imagined (right) data.}
  \label{strategies}
  \vspace{+.5cm}
\end{figure}

Our results are labeled as \enquote{real} if the abstract behavior model was trained on data from actual interactions with the environment and \enquote{imagined} if trained from rollouts within the probabilistic world model. As a first point of comparison, we show the accuracy results from a decision tree in Table \ref{results} at both the overall outcome and at time-varying strategy level for the same set of real test data. For both real and imagined data, we find that we can predict the success or a failure of an episode with over $85\%$ accuracy using the simple conditions discovered by the decision tree. Similarly, we find that our identified conditions predict the correct local strategy with over $75\%$ accuracy. Figure \ref{strategies} shows the prediction breakdown by strategy on test data based on training performed with real and imagined data sets. 

\begin{table}[hbt!]
\caption{Accuracy results from the decision tree}
\label{results}
\begin{center}
\begin{tabular}{  lcc }
 \hline
    & Accuracy & Balanced Accuracy\\
    \hline
    Outcome, Real   & $86.92\%$    & $65.86\%$ \\
    Outcome, Imagined & $85.91\%$    & $63.07\%$ \\
    Strategy, Real & $78.36\%$    & $43.62\%$ \\
    Strategy, Imagined  & $75.52\%$    & $37.76\%$ \\
 \hline
\end{tabular}
\end{center}
\end{table}

\begin{figure}
  \centering
  \includegraphics[scale=0.28]{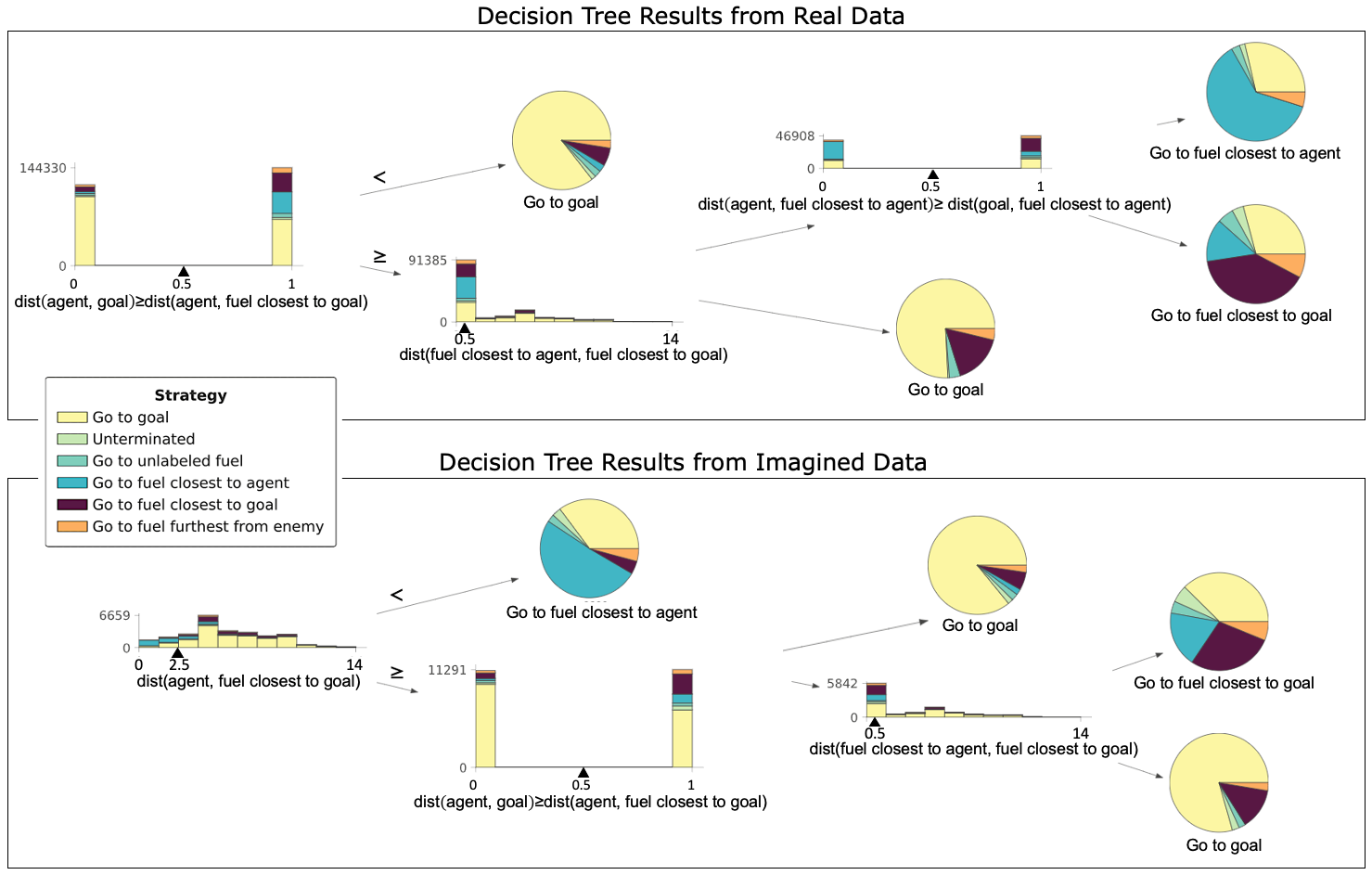}
  \caption{Abstract strategy models learned from the real data (top) and imagined data (bottom)}
  \label{abm}
\end{figure}

To show the experiential conditions identified, we present the decision tree results for both the real and imagined data in Figure \ref{abm}. These models shows the important conditional factors that lead to each of the strategies. For example, the results from the real data show that if the distance between agent and goal is less than the distance between agent and the fuel closest to the goal, then the agent's strategy should be to go directly to the goal. Although the accuracy for the imagined and real trajectory data are similar, there is a variation in condition hierarchy between the two predictions. This provides insight into the differences between the environment and the approximate model, which we plan to explore as part of our future work. 

The potential benefit of abstracting and representing RL behavior in this way is clearly visible from these results. The decision trees are easy for a human to read and interpret, and present the necessary information to pick out the most influential experiential features. Furthermore, they also demonstrate the unbalanced strategy labels which prevents the decision tree from being able to identify the conditions for lesser-seen strategy labels, thus leading to a lower accuracy. This may serve as an indication for us to collect more data and/or re-consider the milestones and the associated strategy labels.

\section{Conclusions and Future Work} \label{futurework}

In this work, we have presented a method for generating abstract behavior models directly from trajectories obtained from an RL system. Trajectories capture experiences, strategies, and outcomes for an RL agent, all of which are important factors that influencing the competency of the system. We process the trajectories directly to obtain experiential feature representations and strategy labels, which are then used in performing condition identification for the behavior of the RL agent. Our objective in distilling trajectory data in this way is to eventually build a competency-communicating RL agent that can maintain user trust when deployed in the real world. The preliminary results shown provide a promising step in that direction. The ability to apply this method to imagined datasets in model-based RL is critical because it reduces the number of interactions required between the trained agent and environment. Additionally, using imagined datasets can allow us to identify the necessary conditions for mission success before deploying an RL agent in a new setting.

There are a multitude of ways this work can be improved upon and expanded. First, the process of identifying strategies can be automated through techniques like clustering and temporal logic inference, thus reducing human effort for more complex RL problems. Furthermore, while we focused on experiential features that are directly identified from the observed states, there are likely to be experiential features that are not explicitly captured in the state representation but instead can be extracted internally from the world model. Similarly, there may be ways to automate the milestone extraction process by directly using the reward function. Extensions of this work may also explore how the identified conditions and strategies may be used to generate an interpretable policy. Additionally, a more concrete analysis on the differences in results between the real and imagined results may be performed to gather insight into the trained world model and how/why it varies from the environment. Finally, our future work includes building upon these abstract behavior models with the aim to eventually develop a competency-aware RL system that can effectively communicate its capabilities and limitations in the real world.

\section*{Acknowledgments}
The authors thank Michael Crystal, Luke Burks, and Mitchell Hebert for their extensive reviews, as well as the entire ALPACA team for their collaboration. This material is based upon work supported by the Defense Advanced Research Projects Agency (DARPA) under Contract No. HR001120C0032. Any opinions, findings and conclusions or recommendations expressed in this material are those of the author(s) and do not necessarily reflect the views of DARPA.

\bibliographystyle{plainnat}
\bibliography{references}

\begin{thebibliography}{23}
\providecommand{\natexlab}[1]{#1}
\providecommand{\url}[1]{\texttt{#1}}
\expandafter\ifx\csname urlstyle\endcsname\relax
  \providecommand{\doi}[1]{doi: #1}\else
  \providecommand{\doi}{doi: \begingroup \urlstyle{rm}\Url}\fi

\bibitem[Chua et~al.(2018)Chua, Calandra, McAllister, and Levine]{chua2018deep}
Kurtland Chua, Roberto Calandra, Rowan McAllister, and Sergey Levine.
\newblock Deep reinforcement learning in a handful of trials using
  probabilistic dynamics models.
\newblock \emph{NIPS}, 2018.

\bibitem[Coppens et~al.(2019)Coppens, Efthymiadis, Lenaerts, Now{\'e}, Miller,
  Weber, and Magazzeni]{coppens2019distilling}
Youri Coppens, Kyriakos Efthymiadis, Tom Lenaerts, Ann Now{\'e}, Tim Miller,
  Rosina Weber, and Daniele Magazzeni.
\newblock Distilling deep reinforcement learning policies in soft decision
  trees.
\newblock \emph{IJCAI Explainable Artificial Intelligence Workshop}, 2019.

\bibitem[Dulac-Arnold et~al.(2019)Dulac-Arnold, Mankowitz, and
  Hester]{dulacarnold2019challenges}
Gabriel Dulac-Arnold, Daniel Mankowitz, and Todd Hester.
\newblock Challenges of real-world reinforcement learning.
\newblock \emph{arXiv preprint arXiv:1904.12901}, 2019.

\bibitem[Ebert et~al.(2018)Ebert, Finn, Dasari, Xie, Lee, and
  Levine]{ebert2018visual}
Frederik Ebert, Chelsea Finn, Sudeep Dasari, Annie Xie, Alex Lee, and Sergey
  Levine.
\newblock Visual foresight: Model-based deep reinforcement learning for
  vision-based robotic control.
\newblock \emph{arXiv preprint arXiv:1812.00568}, 2018.

\bibitem[Finn and Levine(2017)]{Finn_2017}
Chelsea Finn and Sergey Levine.
\newblock Deep visual foresight for planning robot motion.
\newblock \emph{ICRA}, 2017.
\newblock \doi{10.1109/icra.2017.7989324}.

\bibitem[Ha and Schmidhuber(2018)]{ha2018worldmodels}
David Ha and J{\"{u}}rgen Schmidhuber.
\newblock World models.
\newblock \emph{NIPS}, 2018.

\bibitem[Hafner et~al.(2018)Hafner, Lillicrap, Fischer, Villegas, Ha, Lee, and
  Davidson]{hafner2018learning}
Danijar Hafner, Timothy Lillicrap, Ian Fischer, Ruben Villegas, David Ha,
  Honglak Lee, and James Davidson.
\newblock Learning latent dynamics for planning from pixels.
\newblock \emph{arXiv preprint arXiv:1811.04551}, 2018.

\bibitem[Hayes and Shah(2017)]{hayes2017improving}
Bradley Hayes and Julie~A. Shah.
\newblock Improving robot controller transparency through autonomous policy
  explanation.
\newblock \emph{ACM/IEEE HRI}, 2017.

\bibitem[Huang et~al.(2018)Huang, Bhatia, Abbeel, and Dragan]{huangtrust}
Sandy~H. Huang, Kush Bhatia, Pieter Abbeel, and Anca~D. Dragan.
\newblock Establishing appropriate trust via critical states.
\newblock \emph{IROS}, 2018.

\bibitem[Lakshminarayanan et~al.(2017)Lakshminarayanan, Pritzel, and
  Blundell]{lakshminarayanan2017simple}
Balaji Lakshminarayanan, Alexander Pritzel, and Charles Blundell.
\newblock Simple and scalable predictive uncertainty estimation using deep
  ensembles.
\newblock \emph{NIPS}, 2017.

\bibitem[Malik et~al.(2019)Malik, Kuleshov, Song, Nemer, Seymour, and
  Ermon]{malik2019calibrated}
Ali Malik, Volodymyr Kuleshov, Jiaming Song, Danny Nemer, Harlan Seymour, and
  Stefano Ermon.
\newblock Calibrated model-based deep reinforcement learning.
\newblock \emph{ICML}, 2019.

\bibitem[Mnih et~al.(2013)Mnih, Kavukcuoglu, Silver, Graves, Antonoglou,
  Wierstra, and Riedmiller]{mnih2013playing}
Volodymyr Mnih, Koray Kavukcuoglu, David Silver, Alex Graves, Ioannis
  Antonoglou, Daan Wierstra, and Martin Riedmiller.
\newblock Playing atari with deep reinforcement learning.
\newblock \emph{NIPS Deep Learning Workshop}, 2013.

\bibitem[Moerland et~al.(2020)Moerland, Broekens, and
  Jonker]{moerl2020modelbased}
Thomas~M. Moerland, Joost Broekens, and Catholijn~M. Jonker.
\newblock Model-based reinforcement learning: A survey.
\newblock \emph{arXiv preprint arXiv:2006.16712}, 2020.

\bibitem[Nagabandi et~al.(2018)Nagabandi, Kahn, Fearing, and
  Levine]{Nagabandi_2018}
Anusha Nagabandi, Gregory Kahn, Ronald~S. Fearing, and Sergey Levine.
\newblock Neural network dynamics for model-based deep reinforcement learning
  with model-free fine-tuning.
\newblock \emph{ICRA}, 2018.
\newblock \doi{10.1109/icra.2018.8463189}.

\bibitem[Polydoros and Nalpantidis(2017)]{MBRLRobotics}
Athanasios Polydoros and Lazaros Nalpantidis.
\newblock Survey of model-based reinforcement learning: Applications on
  robotics.
\newblock \emph{Journal of Intelligent and Robotic Systems}, 86:\penalty0
  153--173, 2017.

\bibitem[Sequeira and Gervasio(2020)]{Sequeira_2020}
Pedro Sequeira and Melinda Gervasio.
\newblock Interestingness elements for explainable reinforcement learning:
  Understanding agents’ capabilities and limitations.
\newblock In \emph{Artificial Intelligence}, 2020.
\newblock \doi{10.1016/j.artint.2020.103367}.

\bibitem[Shu et~al.(2018)Shu, Xiong, and Socher]{shu2017hierarchical}
Tianmin Shu, Caiming Xiong, and Richard Socher.
\newblock Hierarchical and interpretable skill acquisition in multi-task
  reinforcement learning.
\newblock \emph{ICLR}, 2018.

\bibitem[Sutton and Barto(2018)]{Sutton1998}
Richard~S. Sutton and Andrew~G. Barto.
\newblock \emph{Reinforcement Learning: An Introduction}.
\newblock The MIT Press, second edition, 2018.

\bibitem[Tobin et~al.(2017)Tobin, Fong, Ray, Schneider, Zaremba, and
  Abbeel]{TobinFRSZA17}
Joshua Tobin, Rachel Fong, Alex Ray, Jonas Schneider, Wojciech Zaremba, and
  Pieter Abbeel.
\newblock Domain randomization for transferring deep neural networks from
  simulation to the real world.
\newblock \emph{IROS}, 2017.

\bibitem[van~der Waa et~al.(2018)van~der Waa, van Diggelen, van~den Bosch, and
  Neerincx]{waa2018contrastive}
Jasper van~der Waa, Jurriaan van Diggelen, Karel van~den Bosch, and Mark~A.
  Neerincx.
\newblock Contrastive explanations for reinforcement learning in terms of
  expected consequences.
\newblock \emph{IJCAI Explainable Artificial Intelligence Workshop}, 2018.

\bibitem[Verma et~al.(2018)Verma, Murali, Singh, Kohli, and
  Chaudhuri]{verma2018programmatically}
Abhinav Verma, Vijayaraghavan Murali, Rishabh Singh, Pushmeet Kohli, and Swarat
  Chaudhuri.
\newblock Programmatically interpretable reinforcement learning.
\newblock \emph{ICML}, 2018.

\bibitem[Wang et~al.(2019)Wang, Bao, Clavera, Hoang, Wen, Langlois, Zhang,
  Zhang, Abbeel, and Ba]{wang2019benchmarking}
Tingwu Wang, Xuchan Bao, Ignasi Clavera, Jerrick Hoang, Yeming Wen, Eric
  Langlois, Shunshi Zhang, Guodong Zhang, Pieter Abbeel, and Jimmy Ba.
\newblock Benchmarking model-based reinforcement learning.
\newblock \emph{arXiv preprint arXiv:1907.02057}, 2019.

\bibitem[Łukasz Kaiser et~al.(2020)Łukasz Kaiser, Babaeizadeh, Miłos,
  Osiński, Campbell, Czechowski, Erhan, Finn, Kozakowski, Levine, Mohiuddin,
  Sepassi, Tucker, and Michalewski]{kaiser2019modelbased}
Łukasz Kaiser, Mohammad Babaeizadeh, Piotr Miłos, Błażej Osiński, Roy~H
  Campbell, Konrad Czechowski, Dumitru Erhan, Chelsea Finn, Piotr Kozakowski,
  Sergey Levine, Afroz Mohiuddin, Ryan Sepassi, George Tucker, and Henryk
  Michalewski.
\newblock Model based reinforcement learning for {A}tari.
\newblock \emph{ICLR}, 2020.

\end{thebibliography}

\end{document}